%% file: ms.tex
\documentclass[10pt,twocolumn,letterpaper]{article}

\usepackage{style/cvpr}
\usepackage{times}
\usepackage{epsfig}
\usepackage{graphicx}
\usepackage{amsmath}
\usepackage{amssymb}

\usepackage{float}
\usepackage{subcaption}  

\usepackage[breaklinks=true,colorlinks,bookmarks=false]{hyperref}

\cvprfinalcopy 

\def\cvprPaperID{****} 

\begin{document}

\title{ Distance to Center of Mass Encoding for Instance Segmentation }

\author{Thomio Watanabe\\
University of Sao Paulo\\
{\tt\small thomio.watanabe@usp.br}
\and
Denis Wolf\\
University of Sao Paulo\\
{\tt\small denis@icmc.usp.br}
}



\maketitle

\input{src/abstract.tex}

\input{src/intro.tex}
\input{src/related.tex}

\input{src/dcme.tex}

\input{src/experiments.tex}

\input{src/conclusion.tex}
\input{src/acknowledgement.tex}

\newpage
{\small
\bibliographystyle{bib/ieee}
\bibliography{bib/cvpr.bib}
}

\end{document}

%% file: src/abstract.tex
\begin{abstract}
The instance segmentation can be considered an extension of the object detection problem where bounding boxes are replaced by object contours.
Strictly speaking the problem requires to identify each pixel instance and class independently of the artifice used for this mean.
The advantage of instance segmentation over the usual object detection lies in the precise delineation of objects improving object localization.
Additionally, object contours allow the evaluation of partial occlusion with basic image processing algorithms.
This work approaches the instance segmentation problem as an annotation problem and presents a novel technique to encode and decode ground truth annotations.
We propose a mathematical representation of instances that any deep semantic segmentation model can learn and generalize. 
Each individual instance is represented by a center of mass and a field of vectors pointing to it.
This encoding technique has been denominated Distance to Center of Mass Encoding (DCME).
\end{abstract}

%% file: src/intro.tex
\section{Introduction}

The instance segmentation task aims to precisely localize objects in images and it is a relatively new problem in computer vision.
Currently, the main approach to solve the instance segmentation problem is based in on ensemble methods, a combination of multiple convolutional neural networks or multi-stage neural networks where each network/stage is specialized to solve one specific subtask of the problem.
The work presented by Cordts \etal \cite{Cordts2016Cityscapes} groups instance segmentation solutions in three categories according to the order of the subtasks: segmentation + detection, detection + segmentation and simultaneous detection and segmentation.
Differently from most solutions the Distance to Center of Mass Encoding (DCME) does not rely on detections (region proposal methods).
The DCME is a mathematical representation based in linear algebra to indirectly encode instances on ground truth annotations.
Each instance is represented by a center of mass and 2D displacement vectors.
The center of mass is computed considering each pixel as an unitary mass and a displacement vector is computed with the difference between 2 pixels positions in the image.
The displacement vectors are mapped for each pixel and they point towards their instance center of mass.
Encoding instance information in the ground truth annotations is a straight and simple solution to the problem.
This approach adds overhead to extract the instances but most solutions also need tailored post-processing to extract them. 
To the best of our knowledge there is no other work describing similar representation to solve the instance segmentation problem.
To allow scientific reproducibility a demo will be publicly \href{https://github.com/usp-lrm/dcme}{available}.



Following the success of deep learning \cite{lecun2015deep} this work focus on deep semantic segmentation models \cite{garcia2017review} to solve the instance segmentation problem.
In this context, any pixel-wise deep convolutional neural network is considered a segmentation model.
Experiments were performed on SegNet \cite{badrinarayanan2015segnet} and FCN \cite{long2015fully} to verify the generalization capabilities of segmentation models trained with DCME.
These models were selected to cover two different approaches of deep segmentation models.
SegNet uses max-pooling indices to upsample feature maps and FCN uses deconvolution (transposed convolution) filters.

The success of deep learning models are partially due to the existence of large datasets.
In this context, a dataset quality is directly assessed by the amount of data and the annotations quality.
There are only a few datasets for instance segmentation and the most acknowledged ones are PASCAL VOC \cite{everingham2015pascal,bharath2011semantic}, Microsoft COCO \cite{lin2014microsoft}, and Cityscapes \cite{Cordts2016Cityscapes}.
The proposed solution was evaluated on the challenging Cityscapes dataset because of its urban street context and high quality annotations. 
Depending on the dataset the instance segmentation is also being called object segmentation but this paper favors the Cityscapes naming convention.

The DCME was able to separate individual instances independently of their classes. 
But a complete solution for the instance segmentation problem requires to identify the class of each instance.
The solution was based on SegNet/FCN to separate instances and googleNet \cite{szegedy2015going} to classify them.
Cityscapes is an urban street scene dataset but the solution is theoretically independent of the object type and shape.



%% file: src/related.tex
\section{Related work}


Since detection based solutions usually rely on region proposal techniques, Uhrig \etal \cite{uhrig2016pixel} have separated instance segmentation solutions on proposal-based and proposal-free methods.
Region proposal methods localize areas in images that might present objects generating object candidates.


The Multiscale Combinatorial Grouping (MCG) \cite{arbelaez2014multiscale} computes a hierarchical segmentation on images of different scales to combine them into a single multi-scale segmentation hierarchy.
Finally a grouping component rank a list of object proposals.

More recent, the Region Proposal Networks (RPNs) \cite{ren2015faster} is a CNN built to score region proposals and according to the authors it operates in a sliding-window fashion.
This method has been widely used once it allows multi-stage neural networks to share feature maps.
The great advantage of multi-stage neural networks over stacking multiple networks lies on jointly training the different sub-networks.
This improves feature learning once these different parts share feature maps.

\subsection{Proposal-based methods}

Frequently, proposal-based methods make use of multiple networks trained separately or are single networks with multi-stage tasks. 
They often present the best results in computer vision benchmarks but they are also very computational expensive.
As benchmarks do not evaluate computational cost there is no disadvantage on stacking different methods to solve a problem. 
The scientific value of these solutions are directly associated with benchmark scores.
Unless they reduce the computational cost (number of parameters and/or number of operations) architectures with higher scores are more acknowledged by the scientific community.

Simultaneous Detection and Segmentation \cite{hariharan2014simultaneous} uses MCG to find category-independent region proposals.
Then it uses a convolutional neural network to extract features on each region, a support vector machine to score each candidate and finally employs a non-maximum suppression to select  candidates.

A fully convolutional encoder-decoder network was proposed by Yang \etal \cite{yang2016object} to detect contours from all objects in the image an then uses MCG to separate instances.
A similar approach was proposed by Li \etal \cite{li2017instance} where a segmentation network generates saliency masks and contours masks.
With the contours masks the MCG generates object proposals which are compared to the saliency masks to finally generate the object instances.

The multi-task network cascades \cite{dai2016instance} specializes a single CNN to solve 3 different tasks: instances differentiation, masks estimation and objects categorization.
The first stage is based in the RPNs and proposes class-agnostic bounding boxes.
Later network stages are built upon earlier ones, forming a cascade.

Hayder \etal \cite{hayder2017boundary} propose a solution that is more robust to error in the object candidate generation process based on RPNs.
It proposes the object mask network (OMN) with a residual-deconvolution architecture and integrates the OMN in a multi-task cascade framework with several stages.

\subsection{Proposal-free methods}

Usually, proposal-free solutions are bottom-up approaches that directly use segmentation models. 
These methods do not directly represent instances but they learn specific characteristics like contours, depth or position that are helpful to separate/identify instances.
These solutions are diverse but they are highly dependent on post-processing techniques.

A FCN was used by van den Brand \etal \cite{van2016instance} to separate object contours in 4 classes based in their relative position: top, left, bottom and right.
Instance contours are extracted from the FCN output and the flood fill algorithm is used to define instances masks.
Uhrig \etal \cite{uhrig2016pixel} also employ a FCN to generate a combination of per pixel semantic label, depth and direction.
These three outputs are used by a template matching scheme to extract and classify instances.

A discriminative loss function based in a distance metric was proposed by De Brabandere \etal \cite{de2017semantic} to cluster instances pixels.
Another loss function was designed by Romera-Paredes \cite{romera2016recurrent} to train end-to-end a recurrent neural network.
And, according to the authors, the solution is inspired on how humans count elements in a scene. 
This work assumes the instances belong to a single class.

Zhang \etal \cite{zhang2016instance, zhang2015monocular} model the problem as a Markov Random Field.
These works propose energy functions to train a network with image patches of different scales.
They also propose some heuristics to extract the instances.
These works can only detect instances from a single class and have a limit of 9 instances per image.

Bai and Urtasun \cite{bai2017deep} employ a cascade neural network to predict a watershed energy landscape such that each basin corresponds to a single instance.
The final solution generates a watershed transform energy map representation which is cut at a fixed threshold to yield the final predictions.
The proposal-free network \cite{liang2015proposal} also specializes a convolutional neural network to predict pixel-level semantic labels, the number of instances per class and an instance location vector for each pixel. 

Arnab and Torr \cite{arnab2017pixelwise} utilize several methods to generate objects instances over object detections.
Its final representation is a segmentation map which assigns an object class and instance label for each pixel.
Liu \etal \cite{liu2017sgn} also use a sequence of neural networks to learn breakpoints that form line segments which are then combined to generate instances.

\subsection{Contribution}

This work describes a novel mathematical representation for object instance annotations that deep semantic segmentation models are able to learn and generalize, including this solution among the proposal-free methods.
The approach consists on transforming the annotations to 2D displacement vectors that point towards instances centers of mass.
Encoding and decoding steps are proposed to generate the representations and extract information.

Arnab and Torr \cite{arnab2017pixelwise} also encode instance information in each pixel but this representation is inferred from bounding box annotations and semantic segmentation annotations while DCME indirectly encode instance information on ground truth training annotations.

The work introduced by Uhrig \etal \cite{uhrig2016pixel} and latter extended by Levinkov \etal \cite{levinkov2017joint} also compute directions towards instance centers.
It divides a turn in 8 regions of 45 degrees and each pixel is classified in one of these eight regions according to its relative position to the instance center.
Each one of these 8 regions represent a class inferred by softmax. 
Differently from Uhrig \etal \cite{uhrig2016pixel}, the 2D displacement vectors also encode magnitude information and are represented by two real numbers corresponding to the vector components in each axis. 
Their work also uses pixel-level semantic labels and depth information to train their models. 

This work shares several similarities with Bai and Urtasun \cite{bai2017deep}.
The watershed transform energy map representation is similar to DCME magnitude map, although DCME is not limited to 16 quantization levels.
The DCME aims to generate a 2D displacement vector as the segmentation model output.
While Bai and Urtasun \cite{bai2017deep} employ direction vectors without distance information as an intermediate representation to generate the watershed transform energy map.
The energy map is directly used to extract instances therefore their solution does not solve partial occlusion.

Due to the high scores achieved by proposal-based methods in benchmark datasets one may think this is the best approach to solve the problem, however, some questions stay with no answer. 
What are the most important subtasks to solve the problem?
What is the order among these subtasks?
How these substaks should interact with each other?
It is hard to answer these questions because convolutional neural networks automatically extract the most important features to solve the problem and most of the works in the area are developed empirically.
Although proposal-free methods usually present smaller scores, mathematical models for instance segmentation annotations seems to fit better how deep learning models are currently being employed.

%% file: src/dcme.tex
\section{Distance to center of mass encoding}
The distance to center of mass encoding (DCME) aims to generate instances representations that deep segmentation models are able to learn and generalize.
This work proposes a model to represent instances and algorithms to encode and decode the information.
To create meaningful instance segmentation representation, some points must be addressed:
\begin{itemize}
    \item images have a variable number of instances;
    \item overlapping regions between instances are critical regions;
    \item a single instance may have several disjoint regions (partial occlusion);
\end{itemize}



Differently from the classification and segmentation problem where the number of classes is fixed, the images have a variable number of instances.
Several solutions have a tendency to group together different instances with overlapping regions and 
some instance segmentation solutions are not able to resolve the partial occlusion \cite{bai2017deep, van2016instance}.

The approach to solve these three problems was to generate a representation that could indirectly encode instance information for each pixel.
The pixels are represented by 2D displacement vectors pointing to an instance center of mass (CM). 
Each instance is represented by a single center of mass and the 2D displacement vectors that point to it.
A vector magnitude is the distance between a pixel and its instance center of mass and a vector direction points to the center of mass.
The vectors are represented by 2D components in a Cartesian coordinate system with origin on the image top left corner.
Background pixels that does not represent instances have zero components.

Theoretically, there is no limit for the number of centers of mass and there is no limit for the distance vector magnitude.
Each pixel may become a center of mass, a scene may have any number of instances and these instances have no size restrictions.
However, in practice the number of instances is limited by the image resolution.
The maximum number of instances is half the resolution (full HD ${1920 \times 1080} \div 2 = 1,036,800$ instances), when the other half would be 2D displacement vectors pointing to a single center of mass.
This happens because a center of mass needs at least one displacement vector pointing to it.

As explained in other works \cite{bai2017deep,levinkov2017joint,uhrig2016pixel}, the direction information helps separating overlapping regions since vectors in these areas may have components pointing to opposite directions.
Finally, when each pixel carries information about the instance it belongs to, partial occlusions are background pixels or pixels that belong to other instance.


\subsection{Encoding}
\begin{figure}[H]
    \centering
    \includegraphics[width=0.75\linewidth]{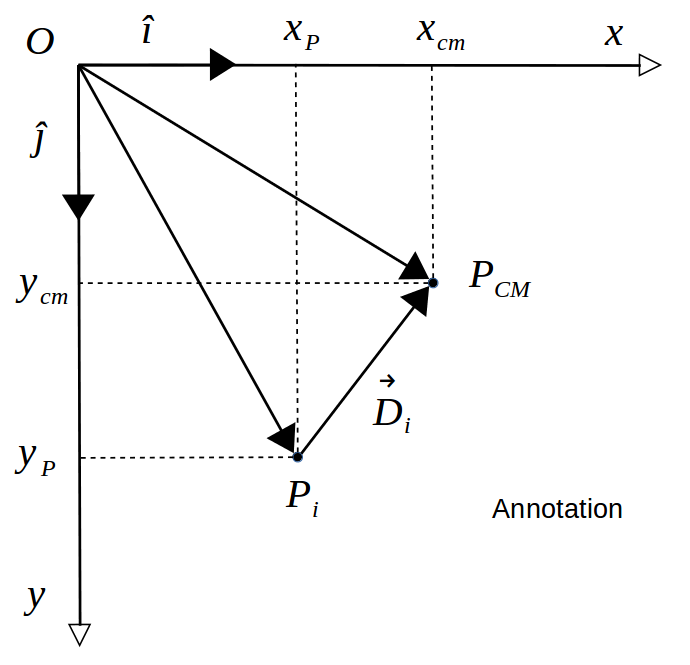}
    \caption{ Displacement vector $\textbf{D}_{i}$ between points $P_{i}$ and $P_{CM}$. Each point $P_{i}$ of instance $I$ has an associated vector $\textbf{D}_{i}$ that points to $P_{CM}$. }
    \label{fig:dcme_encoding}
\end{figure}

Two steps are required to transform the ground truth annotations according to the DCME.
The first step is to find the center of mass position for each object instance, Eq. \ref{eq:center_of_mass}.
The second step computes the displacement vector for each pixel position from each object instance, Eq. \ref{eq:displacement_vector}.
Each pixel position is relative to the coordinate system origin in the annotation top left corner and the $y$ axis points downward, according to Figure \ref{fig:dcme_encoding}.


\begin{itemize}
    \item Step 1 - find center of mass $P_{CM}$ for each instance $I$:
    \begin{subequations}
        \begin{align}
            P_{CM} &= \left( x_{cm}, y_{cm} \right)\\
            x_{cm} &= mean(X); \quad \{X | \forall x \in I\}\\
            y_{cm} &= mean(Y); \quad \{Y | \forall y \in I\}
        \end{align}
    \label{eq:center_of_mass}
    \end{subequations}

    \item Step 2 - compute displacement vector $\textbf{D}_{i}$ for each pixel $ P_{i} = (x_{p},y_{p})$ from each instance $I$:
    \begin{subequations}
        \begin{align}
        \textbf{D}_{i} &= \textbf{OP}_{CM} - \textbf{OP}_{i} \quad \{\forall P_{i} \in I\}\\
        \textbf{D}_{i} &= d_{x} \textbf{i} + d_{y} \textbf{j}\\
        d_{x} &= x_{cm} - x_{p}\\
        d_{y} &= y_{cm} - y_{p}
        \end{align}
    \label{eq:displacement_vector}
    \end{subequations}
\end{itemize}


The center of mass position is the average of the columns and rows of the pixels belonging to the instance.
The mass is considered uniformly distributed and each pixel has unitary mass.
Therefore, there is no heavier regions pulling the center of mass which tends to be located in regions with higher superficial density of pixels. 

\begin{figure}[H]
    \centering
    \includegraphics[width=0.8\linewidth]{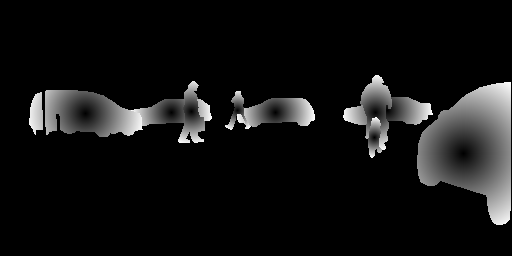}
    \includegraphics[width=0.8\linewidth]{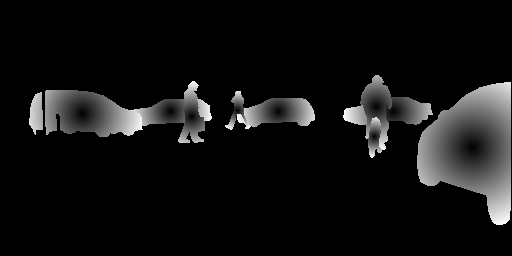}
    \caption{ Magnitude map on Cityscapes ground truth annotation. Top: distance to bounding box centroid. Bottom: distance to center of mass. Intensity is directly proportional to 2D vector magnitude. Left car presents partial occlusion. Rider and bicycle centers of mass are farther than corresponding bounding box centroid. }
    \label{fig:magnitude_map}
\end{figure}

Initially the instance bounding box centroid was used in place of the center of mass.
However, this anchor was very susceptible to outliers, where a few pixels placed far from the instance would greatly affect its position.
The centroid of close instances tend to be very close to each other reducing the decodification effectiveness.
Once the center of mass is located in dense regions it improves separation over close instances, Figure \ref{fig:magnitude_map}.

For convenience the final annotation generated by the DCME will be denominated 2D vector map.
This representation have same spatial dimensions as the input and each pixel is a vector with associated components $d_{x}$ and $d_{y}$.

Once the vector components belong to the real numbers set $\mathbb{R}$ the classification softmax layer from the segmentation models must be replaced by a regression layer.
This is the single modification in the model architecture required by this solution.
The segmentation model will not compute anymore a probability distribution over a fixed number of classes but it will directly infer the displacement vector components. 

\subsection{Decoding}
After training, the segmentation model will be able to generate results similar to the annotations.
The decoding step intends to extract the instances from the vector maps.
The decodification step requires to find the centers of mass which are local \textit{minima} in the magnitude surface.
The magnitude and direction are noisy and in consequence the magnitude surface is irregular with several local \textit{minima}.
Therefore, an heuristic was developed select the most likely centers of mass.

Initially, the center of mass pointed by each 2D vector in evaluated.
Due to the segmentation model output noise the vectors will point to a region close to the instance center of mass.
The centers of mass closer to each other are clustered and only those that a minimal number of vectors are pointing at, are selected.
An instance is the ensemble of all the vectors that point to the region around the center of mass.
In summary, the decoding is composed of the following steps:

\begin{enumerate}
    \item compute center of mass proposals.
    \item cluster close centers of mass $\rightarrow$ distance threshold (DT).
    \item select centers of mass pointed by a minimal number of vectors $\rightarrow$ vote threshold (VT).
    \item select vectors that point to the center of mass region $\rightarrow$ error threshold (ET).
\end{enumerate}

There are three thresholds associated with this solution and a compromise among them is required to achieve good results. 
A high vote threshold tends to discard small instances with small number of pixels pointing to the the center of mass.
Smaller vote thresholds include small instances but also include several false positives.

Large instances have large areas with potential centers of mass.
High error threshold and high distance threshold improve the detection of large instances pixels but small instances that are close to each other will tend to be grouped as a single instance.


%% file: src/experiments.tex
\section{Experiments}

The DCME was evaluated on Cityscapes urban street scene benchmark and dataset for pixel-level and instance-level semantic labeling \cite{Cordts2016Cityscapes}.
The dataset was designed to leverage the understanding of complex traffic scenes and driving scenarios.
Only the fine annotation train set was used to train the models in a single phase fashion and no data augmentation, data balancing or pre-processing technique was applied to improve score results.
However the images and annotations had to be subsampled by a factor of 3 reducing their resolution from (1024, 2048) to (340, 680).
The experiments were performed with the Caffe deep learning framework \cite{jia2014caffe}. 

\subsection{Instance segmentation pipeline}

The DCME can be used to separate instances and it does not encode information about instances classes. 
Currently, to completely solve the instance segmentation problem a model is required to classify each one of the instances.
The classification step can be done by any classification model and it is only required when the dataset presents more than one class.
If the problem has multiple instances of a single class, the classification step would not be required.

Three convolutional neural networks were used to evaluate the DCME.
Two semantic segmentation models, SegNet and FCN-8s to separate instances and the googleNet classification model to classify the extracted instances.
These two segmentation models were chosen to cover the two different approaches for image segmentation thus demonstrating DCME is independent of the segmentation model architecture/paradigm. 
Both have different decoders and additionally the SegNet model uses the max poling indices to upsample feature maps while the FCN uses the deconvolution layer, also known as transposed convolutional layer.
The basic-SegNet model was also evaluated but its results were roughly half of the main SegNet model and they are not presented here.

Both segmentation networks were modified to yield a 2D output with same spatial dimension of the input images/annotations.
These networks perform a regression and their weights were optimized with the mean squared error. 
The SegNet network was trained with a $10^{-8}$ learning rate and a batch size of 15 images/annotations, the FCN-8s network was trained with a $10^{-10}$ learning rate and a batch size of 14 images/annotations.
The decodification thresholds were defined as (DT, VT, ET): DCME-SegNet (10, 50, 15) and DCME-FCN (15, 30, 20).

The googleNet inception-v1 model was used to classify the instances.
To train the classification network, the instances from the fine train set were extracted and resized to (224, 224).
It was trained with the ``poly'' policy, initial learning rate of $10^{-3}$ and batch size of 64 images.
The segmentation and classification models are trained separately and future efforts will be directed to integrate these two phases.


Due to the high computational cost of the segmentation model, the original images and annotations had to be subsampled by a factor of 3 (1024, 2048) to (340, 680).
The subsampling transformation destroys information required to precisely delineate object contours.
Moreover, the final result must be upsampled to the original resolution to be evaluated, adding noise to it.
Therefore, the reduction on the image resolution has a great impact over the final score.
Evaluating the validation set ground truth annotations with a (340, 680) resolution provides only 39.8\% AP.


\subsection{Evaluation}

The Cityscapes instance segmentation problem has 8 classes where 94\% of the fine training set instances belong to 4 classes: car, person, bicycle and rider.
This imbalance is likely a representation of the real scenario on German urban streets rather than a design error.
There is no size limit to detect instances and the high resolution of the images make the Cityscapes very challenging.

Cityscapes evaluate instance segmentation submissions based on variations of the Average Precision (AP).
Two submission were made and 
in both evaluations the same googleNet model was used to classify the instances among the 8 classes.
The DCME-SegNet scores will be published on Cityscapes webpage.
These solutions are compared among each other and other Cityscapes solutions.
Also, some specific modifications in the proposed solution are evaluated.

\begin{figure*}
\begin{center}
	\begin{subfigure}{\textwidth}
        \includegraphics[scale=0.25]{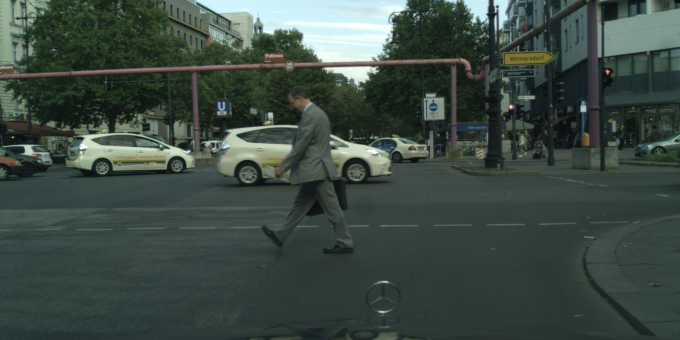}
        \includegraphics[scale=0.25]{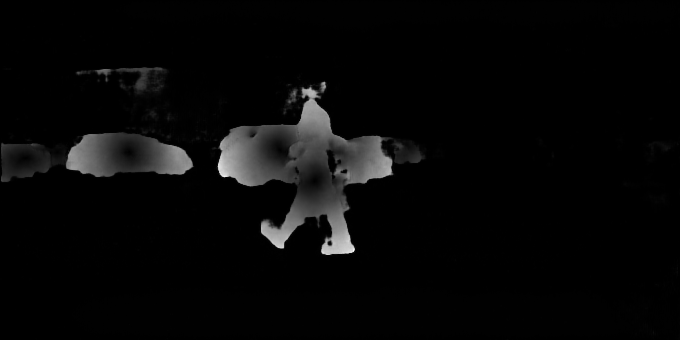}
        \includegraphics[scale=0.25]{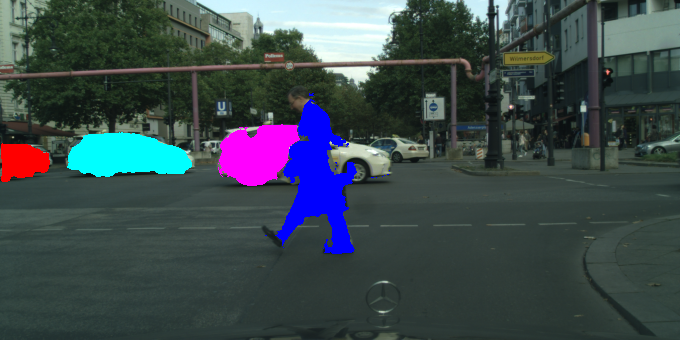}
	\end{subfigure}
	\begin{subfigure}{\textwidth}
        \includegraphics[scale=0.25]{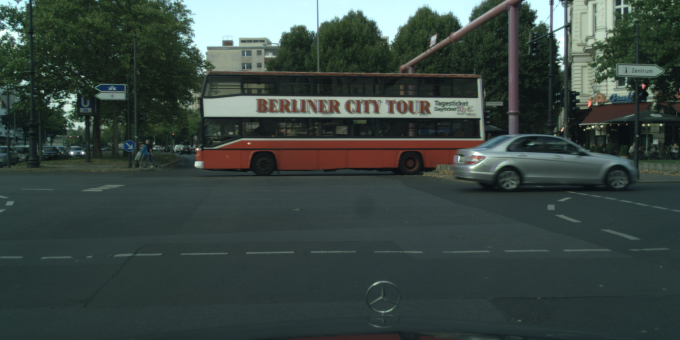}
        \includegraphics[scale=0.25]{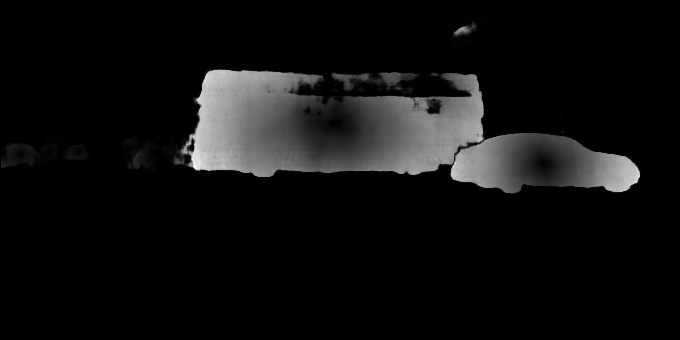}
        \includegraphics[scale=0.25]{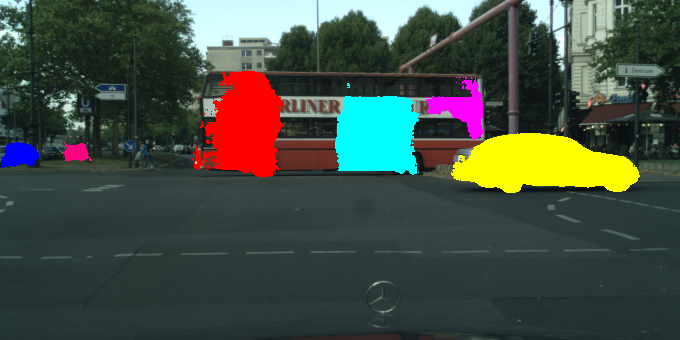}
	\end{subfigure}
	\begin{subfigure}{\textwidth}
        \includegraphics[scale=0.25]{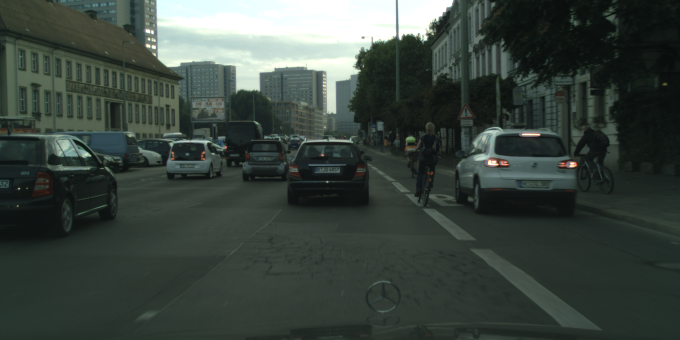}
        \includegraphics[scale=0.25]{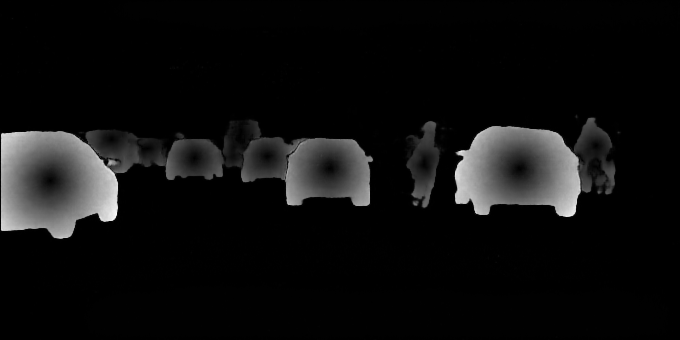}
        \includegraphics[scale=0.25]{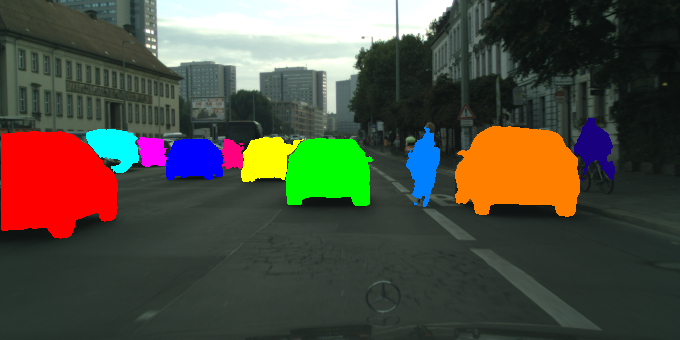}
	\end{subfigure}
	\begin{subfigure}{\textwidth}
        \includegraphics[scale=0.25]{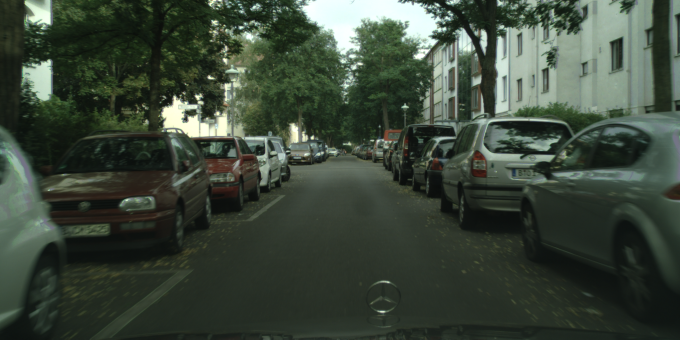}
        \includegraphics[scale=0.25]{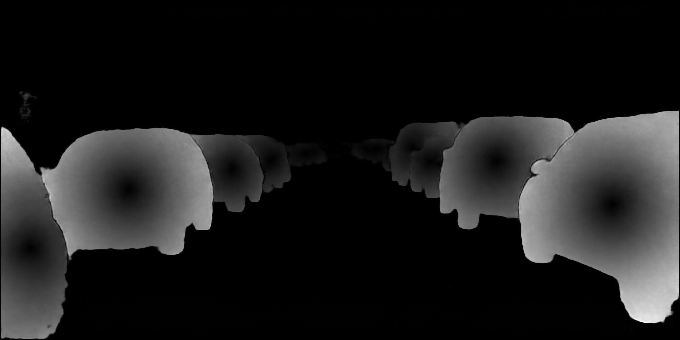}
        \includegraphics[scale=0.25]{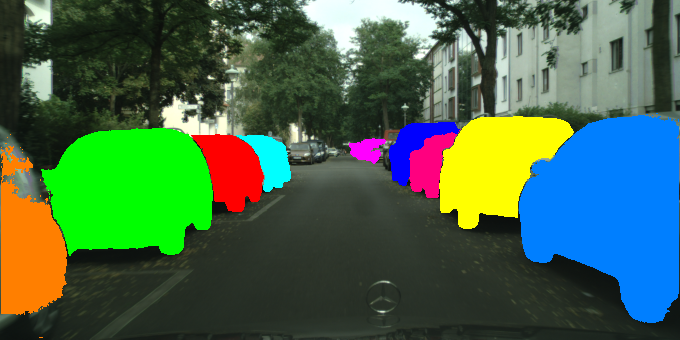}
	\end{subfigure}
	\begin{subfigure}{\textwidth}
        \includegraphics[scale=0.25]{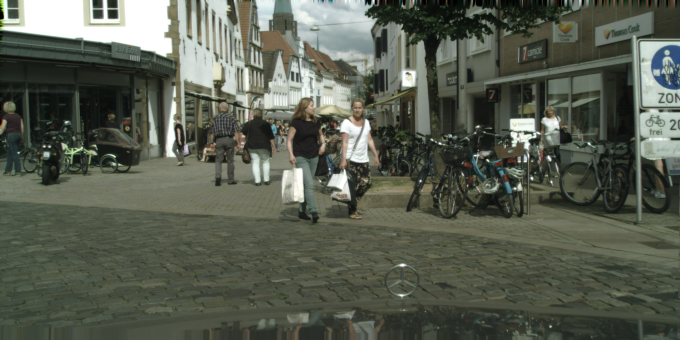}
        \includegraphics[scale=0.25]{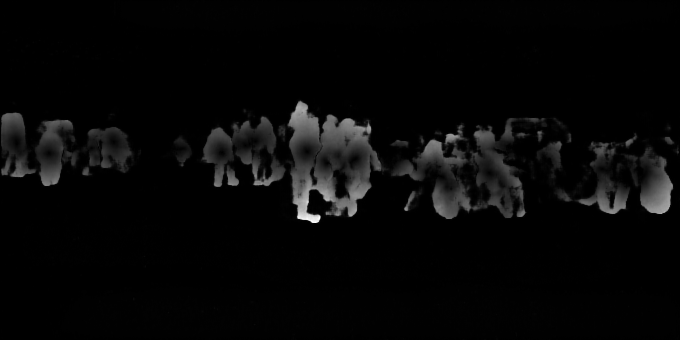}
        \includegraphics[scale=0.25]{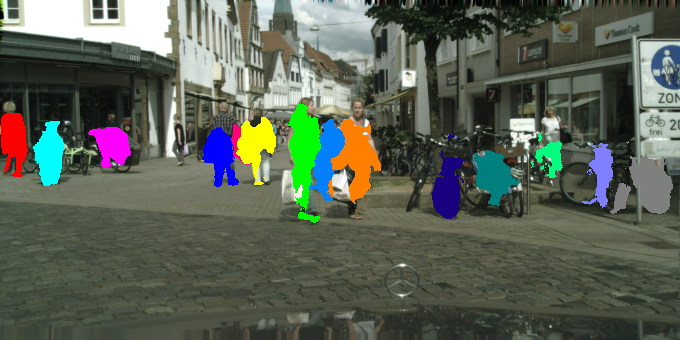}
	\end{subfigure}
\end{center}
\caption{DCME-SegNet results on Cityscapes fine annotations test set. From left to right: input images, 2D vectors magnitude map and output masks. The segmentation model generates 2D vector maps (magnitude and direction) and the masks are generated by decoding these vectors. Different mask colors represent different instances. Instances masks are later separated in classes. }
\label{fig:segnet_output}
\end{figure*}

\begin{figure*}
\begin{center}
	\begin{subfigure}{\textwidth}
        \includegraphics[scale=0.25]{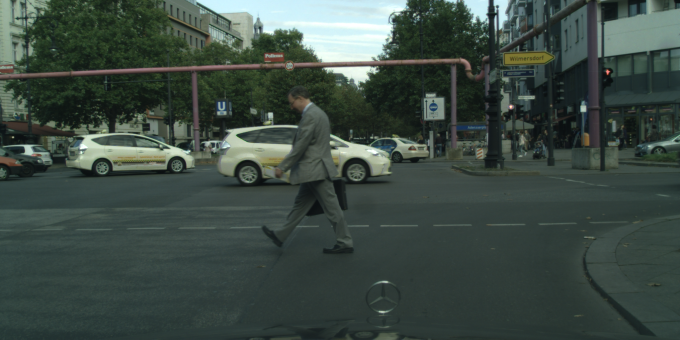}
        \includegraphics[scale=0.25]{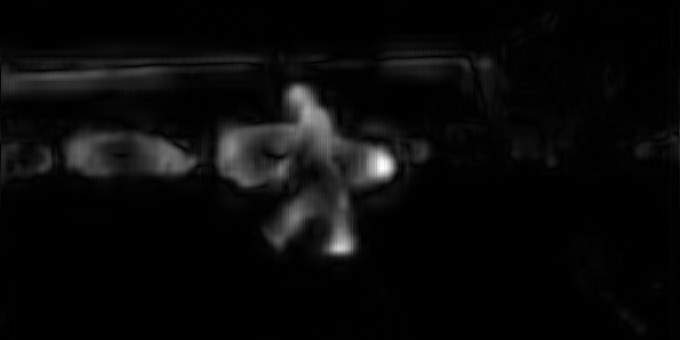}
        \includegraphics[scale=0.25]{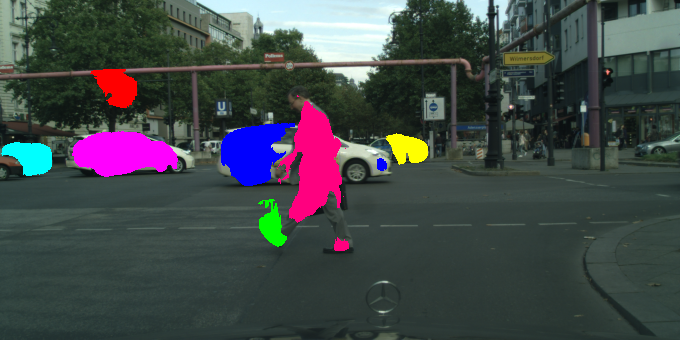}
    \end{subfigure}
    \begin{subfigure}{\textwidth}
        \includegraphics[scale=0.25]{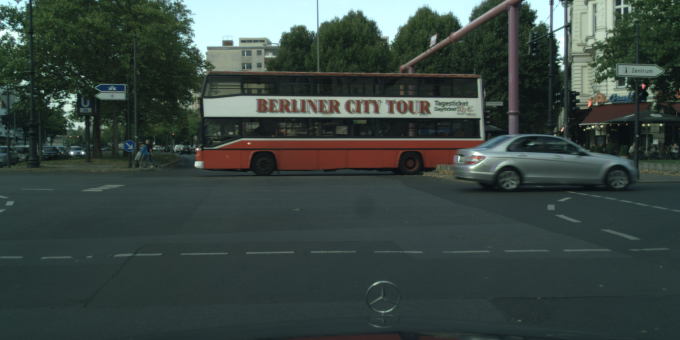}
        \includegraphics[scale=0.25]{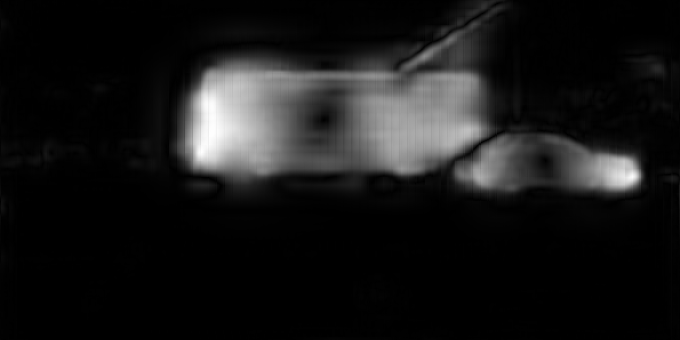}
        \includegraphics[scale=0.25]{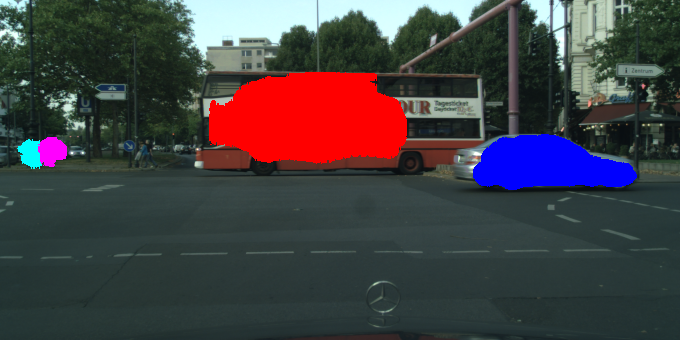}
    \end{subfigure}
    \begin{subfigure}{\textwidth}
        \includegraphics[scale=0.25]{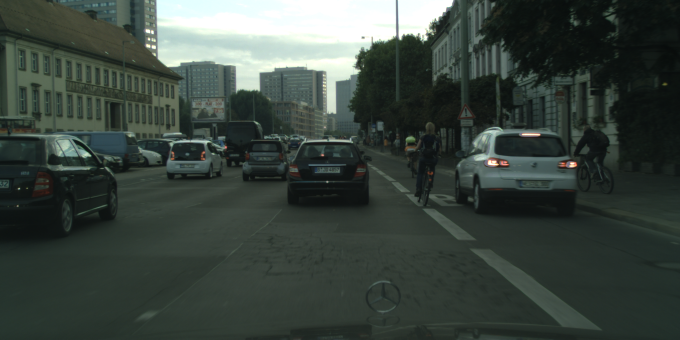}
        \includegraphics[scale=0.25]{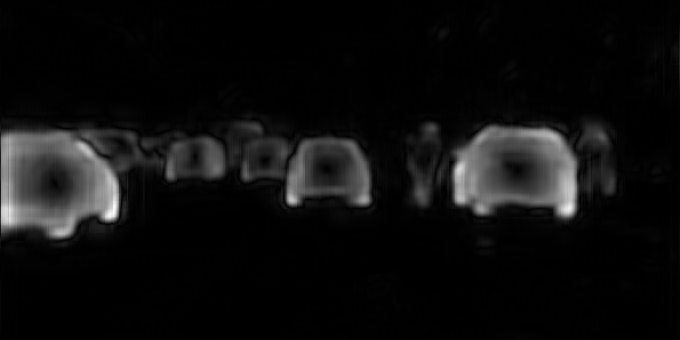}
        \includegraphics[scale=0.25]{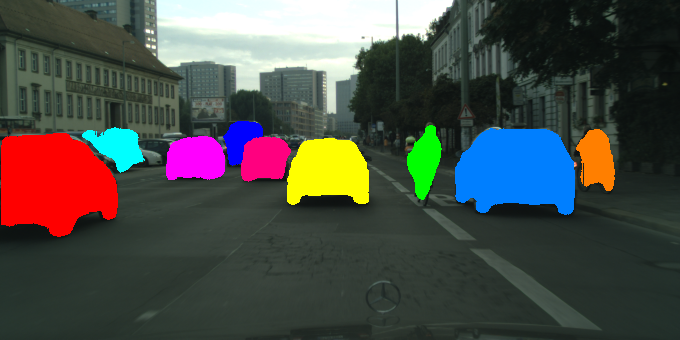}
    \end{subfigure}
    \begin{subfigure}{\textwidth}
        \includegraphics[scale=0.25]{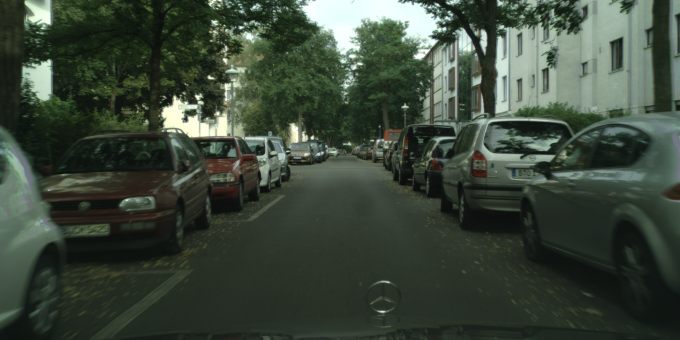}
        \includegraphics[scale=0.25]{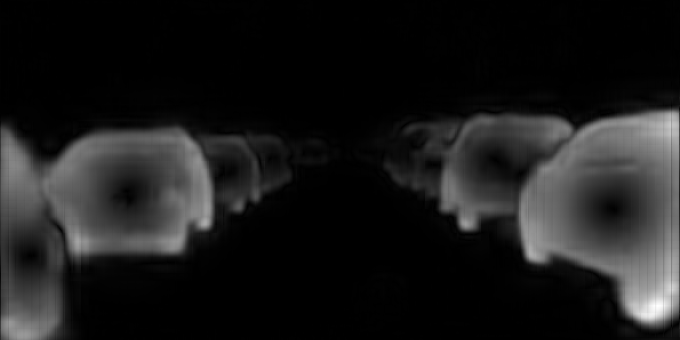}
        \includegraphics[scale=0.25]{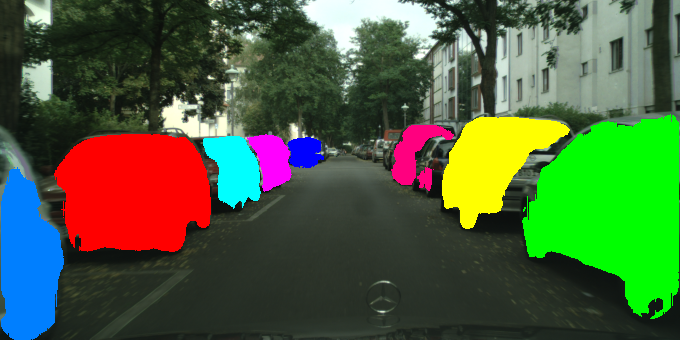}
    \end{subfigure}
    \begin{subfigure}{\textwidth}
        \includegraphics[scale=0.25]{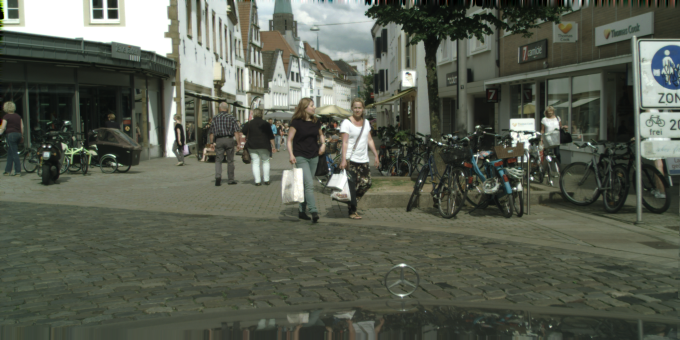}
        \includegraphics[scale=0.25]{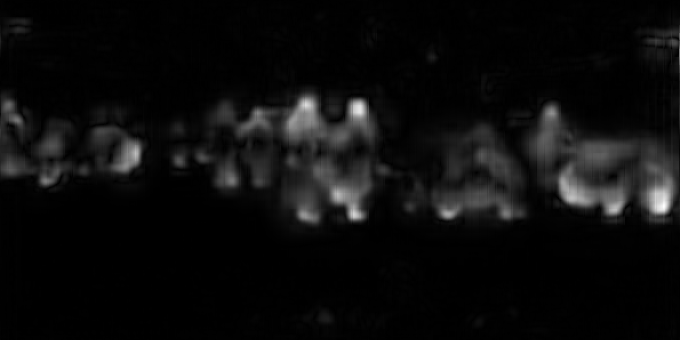}
        \includegraphics[scale=0.25]{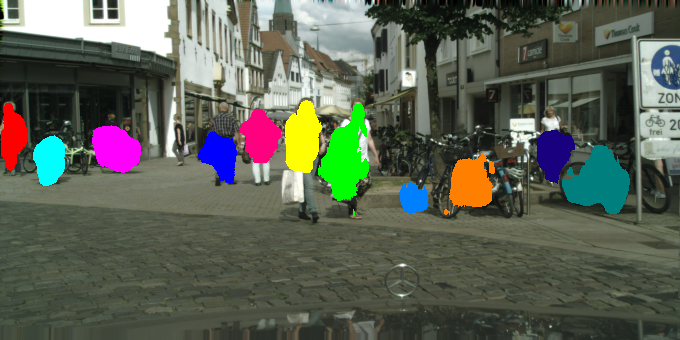}
    \end{subfigure}
\end{center}
\caption{DCME-FCN results on Cityscapes fine annotations test set. From left to right: input images, 2D vectors magnitude map and output masks.  The segmentation model generates 2D vector maps (magnitude and direction) and the masks are generated by decoding these vectors. Different mask colors represent different instances. Instances masks are later separated in classes. }
\label{fig:fcn_output}
\end{figure*}

Figures \ref{fig:segnet_output} and \ref{fig:fcn_output} present the output masks generated by both solutions on the fine annotation test set.
The magnitude maps represent the distance between the pixels and their instance center of mass.
Small instances are not visible in the magnitude maps once the values were normalized by the maximum magnitude and then scaled to fit the 0-255 interval.
Comparing DCME-SegNet and DCME-FCN magnitude maps it is possible to notice that the segmentation model quality has a great impact over the final result.
The DCME-FCN magnitude map has an blurry aspect while the DCME-SegNet presents a smooth intensity gradient.
The DCME-SegNet is able to precisely delineate car instances, clearly representing wing mirrors.
These images demonstrate evidences that the main solution bottleneck is the segmentation model quality.
This includes not only the selection of better segmentation network architectures but also design options like image resolution, data augmentation and data balancing.
These may be viewed as technical problems rather than scientific.

There are only 14 entries on the benchmark indicating that instance segmentation is still an open problem, Table \ref{tab:cityscapes_evaluation}.
Currently the best performing solution, method 1, has 31.95\% AP.
This is still a low score when comparing to the best performing solutions on image classification, object detection and semantic segmentation benchmarks.
Method 1 is proposal-based and methods 12, 13 and 14 are proposal-free.

\begin{table}[H]
\centering
\caption{ DCME evaluation on Cityscapes: SegNet + GoogleNet and FCN-8s + GoogleNet }
\label{tab:cityscapes_evaluation}
\begin{tabular}{ c r r r r }
    \hline
    Method      &    AP & AP50\%& AP100m& AP50m \\
    \hline
    1           & 31.95 & 58.11 & 45.77 & 49.46 \\
    ...         &       & ...   &       &       \\
    12          &  8.89 & 21.14 & 15.26 & 16.71 \\
    13          &  4.55 & 12.90 &  7.72 & 10.26 \\
    DCME-SegNet &  3.77 &  7.73 &  6.62 &  9.47 \\
    DCME-FCN    &  3.47 &  7.86 &  6.12 &  8.61 \\
    14          &  2.27 &  3.65 &  3.88 &  4.87 \\
    \hline
\end{tabular}
\end{table}

\begin{itemize}
\item Method 1 - Mask R-CNN \cite{he2017mask}
\item Method 12 - Pixel-level Encoding for Instance Segmentation \cite{uhrig2016pixel}
\item Method 13 - R-CNN + MCG convex hull \cite{Cordts2016Cityscapes}
\item Method 14 - Instance-level Segmentation of Vehicles by Deep Contours \cite{van2016instance}
\end{itemize}

The DCME-SegNet AP is a little higher than the DCME-FCN, this is likely due the fact SegNet is usually more precise than FCN, presenting higher scores in segmentation benchmarks.
The AP100m and AP50m scores compute the AP over objects within 100m and 50m range.
In both submissions these two scores are close to those presented by method 13 while the AP50\% is roughly the half.
This indicates the DCME-decoding tends to discard small instances. 

Tables \ref{tab:cityscapes_evaluation_segnet} and \ref{tab:cityscapes_evaluation_fcn} present per class scores from both submissions.
The DCME-SegNet has higher scores than the method 13 for the classes person, rider and car.
The low score on other classes may be explained by the high data imbalance of the dataset.
On the fine train set, all the instances that belong to the classes truck, bus, train and motorcycle represent only 3.65\% of all instances.
Applying artificial data balancing techniques may increase score over these minority classes.

\begin{table}[H]
\centering
\caption{ DCME-SegNet + GoogleNet per class evaluation on Cityscapes. }
\label{tab:cityscapes_evaluation_segnet}
\begin{tabular}{ c r r r r }
    \hline
    Class      &    AP & AP50\%& AP100m& AP50m \\
    \hline
    person     &  1.77 &  5.86 &  3.75 &  4.06 \\
    rider      &  0.71 &  3.33 &  1.29 &  1.38 \\
    car        & 15.53 & 25.65 & 26.60 & 35.54 \\
    truck      &  2.00 &  4.02 &  3.58 &  5.12 \\
    bus        &  4.30 &  8.30 &  8.14 & 14.66 \\
    train      &  4.57 &  9.98 &  7.73 & 12.86 \\
    motorcycle &  0.93 &  3.39 &  1.30 &  1.50 \\
    bicycle    &  0.33 &  1.35 &  0.58 &  0.61 \\
    \hline
\end{tabular}
\end{table}

\begin{table}[H]
\centering
\caption{ DCME-FCN + GoogleNet per class evaluation on Cityscapes. }
\label{tab:cityscapes_evaluation_fcn}
\begin{tabular}{ c r r r r }
    \hline
    Class      &    AP & AP50\%& AP100m& AP50m \\
    \hline
    person     &  1.33 &  5.07 &  2.84 &  3.07 \\
    rider      &  0.55 &  2.57 &  0.97 &  1.04 \\
    car        & 14.19 & 24.72 & 24.33 & 33.28 \\
    truck      &  1.24 &  3.16 &  2.25 &  2.94 \\
    bus        &  5.72 & 12.62 & 10.84 & 19.18 \\
    train      &  3.04 &  8.02 &  5.13 &  6.42 \\
    motorcycle &  1.29 &  5.05 &  1.88 &  2.17 \\
    bicycle    &  0.40 &  1.64 &  0.74 &  0.78 \\
    \hline
\end{tabular}
\end{table}


Also, the AP from the validation set was around 6.1\% for the DCME-SegNet and 6.2\% for the DCME-FCN, therefore any technique to reduce the generalization error like data augmentation could effectively improve the final score.
Although this solution presents low scores, the main evaluation purpose was to confirm that deep segmentation models are able to learn and generalize the DCME representation.
Both SegNet and FCN-8s were able to separate instance for all eight classes presenting results compatible with other instance segmentation solutions.

A different decoding was evaluated replacing its fourth step by the watershed algorithm.
The results were slightly worse for the DCME-SegNet scoring 5.7\% AP on the validation data set.
But the results were even worse for the DCME-FCN which scored 1.8\% AP. 
In both cases the watershed could not detect regions close to contours.
The watershed is computed over the magnitude map and it does not make use of direction information.
Therefore the watershed worse performance may be explained because it uses less information than the proposed decoding.



%% file: src/conclusion.tex
\section{Conclusion}
This work presents a novel encoding technique to represent object instances in images.
The representation is capable to separate object instances independently of their classes.
Deep semantic segmentation models with minor modifications were able to learn this mathematical representation.
The model was evaluated in the context of instance segmentation problem on Cityscapes dataset presenting promising results.
Future work will investigate solutions to integrate regression and classification in a single network and improvements over the DCME decoding.


%% file: src/acknowledgement.tex
\subsection*{Acknowledgement}

This work was funded by Sao Paulo Research Foundation (FAPESP) project: \#2015/26293-0.

%% file: ms.bbl
\begin{thebibliography}{10}\itemsep=-1pt

\bibitem{arbelaez2014multiscale}
P.~Arbel{\'a}ez, J.~Pont-Tuset, J.~T. Barron, F.~Marques, and J.~Malik.
\newblock Multiscale combinatorial grouping.
\newblock In {\em Proceedings of the IEEE conference on computer vision and
  pattern recognition}, pages 328--335, 2014.

\bibitem{arnab2017pixelwise}
A.~Arnab and P.~H.~S. Torr.
\newblock Pixelwise instance segmentation with a dynamically instantiated
  network.
\newblock In {\em 2017 IEEE Conference on Computer Vision and Pattern
  Recognition (CVPR)}, pages 879--888, July 2017.

\bibitem{badrinarayanan2015segnet}
V.~Badrinarayanan, A.~Kendall, and R.~Cipolla.
\newblock Segnet: A deep convolutional encoder-decoder architecture for image
  segmentation.
\newblock In {\em IEEE Transactions on Pattern Analysis and Machine
  Intelligence}. IEEE, 2017.

\bibitem{bai2017deep}
M.~Bai and R.~Urtasun.
\newblock Deep watershed transform for instance segmentation.
\newblock In {\em 2017 IEEE Conference on Computer Vision and Pattern
  Recognition (CVPR)}, pages 2858--2866, July 2017.

\bibitem{Cordts2016Cityscapes}
M.~Cordts, M.~Omran, S.~Ramos, T.~Rehfeld, M.~Enzweiler, R.~Benenson,
  U.~Franke, S.~Roth, and B.~Schiele.
\newblock The cityscapes dataset for semantic urban scene understanding.
\newblock In {\em Proceedings of the IEEE Conference on Computer Vision and
  Pattern Recognition (CVPR)}, 2016.

\bibitem{dai2016instance}
J.~Dai, K.~He, and J.~Sun.
\newblock Instance-aware semantic segmentation via multi-task network cascades.
\newblock In {\em Proceedings of the IEEE Conference on Computer Vision and
  Pattern Recognition}, pages 3150--3158, 2016.

\bibitem{de2017semantic}
B.~De~Brabandere, D.~Neven, and L.~Van~Gool.
\newblock Semantic instance segmentation with a discriminative loss function.
\newblock {\em arXiv preprint arXiv:1708.02551}, 2017.

\bibitem{everingham2015pascal}
M.~Everingham, S.~A. Eslami, L.~Van~Gool, C.~K. Williams, J.~Winn, and
  A.~Zisserman.
\newblock The pascal visual object classes challenge: A retrospective.
\newblock {\em International journal of computer vision}, 111(1):98--136, 2015.

\bibitem{garcia2017review}
A.~Garcia-Garcia, S.~Orts-Escolano, S.~Oprea, V.~Villena-Martinez, and
  J.~Garcia-Rodriguez.
\newblock A review on deep learning techniques applied to semantic
  segmentation.
\newblock {\em arXiv preprint arXiv:1704.06857}, 2017.

\bibitem{bharath2011semantic}
B.~Hariharan, P.~Arbelaez, L.~Bourdev, S.~Maji, and J.~Malik.
\newblock Semantic contours from inverse detectors.
\newblock In {\em International Conference on Computer Vision (ICCV)}, 2011.

\bibitem{hariharan2014simultaneous}
B.~Hariharan, P.~Arbel{\'a}ez, R.~Girshick, and J.~Malik.
\newblock Simultaneous detection and segmentation.
\newblock In {\em European Conference on Computer Vision}, pages 297--312.
  Springer, 2014.

\bibitem{hayder2017boundary}
Z.~Hayder, X.~He, and M.~Salzmann.
\newblock Boundary-aware instance segmentation.
\newblock In {\em Conference on Computer Vision and Pattern Recognition
  (CVPR)}, 2017.

\bibitem{he2017mask}
K.~He, G.~Gkioxari, P.~Doll{\'a}r, and R.~Girshick.
\newblock Mask r-cnn.
\newblock {\em arXiv preprint arXiv:1703.06870}, 2017.

\bibitem{jia2014caffe}
Y.~Jia, E.~Shelhamer, J.~Donahue, S.~Karayev, J.~Long, R.~Girshick,
  S.~Guadarrama, and T.~Darrell.
\newblock Caffe: Convolutional architecture for fast feature embedding.
\newblock In {\em Proceedings of the 22nd ACM international conference on
  Multimedia}, pages 675--678. ACM, 2014.

\bibitem{lecun2015deep}
Y.~LeCun, Y.~Bengio, and G.~Hinton.
\newblock Deep learning.
\newblock {\em Nature}, 521(7553):436--444, 2015.

\bibitem{levinkov2017joint}
E.~Levinkov, J.~Uhrig, S.~Tang, M.~Omran, E.~Insafutdinov, A.~Kirillov,
  C.~Rother, T.~Brox, B.~Schiele, and B.~Andres.
\newblock Joint graph decomposition \& node labeling: Problem, algorithms,
  applications.
\newblock In {\em IEEE Conference on Computer Vision and Pattern Recognition
  (CVPR)}, 2017.

\bibitem{li2017instance}
G.~Li, Y.~Xie, L.~Lin, and Y.~Yu.
\newblock Instance-level salient object segmentation.
\newblock In {\em 2017 IEEE Conference on Computer Vision and Pattern
  Recognition (CVPR)}, pages 247--256, July 2017.

\bibitem{liang2015proposal}
X.~Liang, Y.~Wei, X.~Shen, J.~Yang, L.~Lin, and S.~Yan.
\newblock Proposal-free network for instance-level object segmentation.
\newblock {\em arXiv preprint arXiv:1509.02636}, 2015.

\bibitem{lin2014microsoft}
T.-Y. Lin, M.~Maire, S.~Belongie, J.~Hays, P.~Perona, D.~Ramanan,
  P.~Doll{\'a}r, and C.~L. Zitnick.
\newblock Microsoft coco: Common objects in context.
\newblock In {\em European conference on computer vision}, pages 740--755.
  Springer, 2014.

\bibitem{liu2017sgn}
S.~Liu, J.~Jia, S.~Fidler, and R.~Urtasun.
\newblock Sgn: Sequential grouping networks for instance segmentation.
\newblock In {\em The IEEE International Conference on Computer Vision (ICCV)},
  Oct 2017.

\bibitem{long2015fully}
J.~Long, E.~Shelhamer, and T.~Darrell.
\newblock Fully convolutional networks for semantic segmentation.
\newblock In {\em The IEEE Conference on Computer Vision and Pattern
  Recognition (CVPR)}, June 2015.

\bibitem{ren2015faster}
S.~Ren, K.~He, R.~Girshick, and J.~Sun.
\newblock Faster r-cnn: Towards real-time object detection with region proposal
  networks.
\newblock In {\em Advances in neural information processing systems}, pages
  91--99, 2015.

\bibitem{romera2016recurrent}
B.~Romera-Paredes and P.~H.~S. Torr.
\newblock Recurrent instance segmentation.
\newblock In {\em European Conference on Computer Vision}, pages 312--329.
  Springer, 2016.

\bibitem{szegedy2015going}
C.~Szegedy, W.~Liu, Y.~Jia, P.~Sermanet, S.~Reed, D.~Anguelov, D.~Erhan,
  V.~Vanhoucke, and A.~Rabinovich.
\newblock Going deeper with convolutions.
\newblock In {\em Proceedings of the IEEE Conference on Computer Vision and
  Pattern Recognition}, pages 1--9, 2015.

\bibitem{uhrig2016pixel}
J.~Uhrig, M.~Cordts, U.~Franke, and T.~Brox.
\newblock Pixel-level encoding and depth layering for instance-level semantic
  labeling.
\newblock In {\em German Conference on Pattern Recognition}, pages 14--25.
  Springer, 2016.

\bibitem{van2016instance}
J.~van~den Brand, M.~Ochs, and R.~Mester.
\newblock Instance-level segmentation of vehicles by deep contours.
\newblock In {\em Asian Conference on Computer Vision}, pages 477--492.
  Springer, 2016.

\bibitem{yang2016object}
J.~Yang, B.~Price, S.~Cohen, H.~Lee, and M.-H. Yang.
\newblock Object contour detection with a fully convolutional encoder-decoder
  network.
\newblock In {\em Proceedings of the IEEE Conference on Computer Vision and
  Pattern Recognition}, pages 193--202, 2016.

\bibitem{zhang2016instance}
Z.~Zhang, S.~Fidler, and R.~Urtasun.
\newblock Instance-level segmentation for autonomous driving with deep densely
  connected mrfs.
\newblock In {\em Proceedings of the IEEE Conference on Computer Vision and
  Pattern Recognition}, pages 669--677, 2016.

\bibitem{zhang2015monocular}
Z.~Zhang, A.~G. Schwing, S.~Fidler, and R.~Urtasun.
\newblock Monocular object instance segmentation and depth ordering with cnns.
\newblock In {\em Proceedings of the IEEE International Conference on Computer
  Vision}, pages 2614--2622, 2015.

\end{thebibliography}
